%% file: main.tex
\documentclass{article} 
\usepackage{iclr2021_conference,times}

\input{math_commands.tex}

\usepackage{hyperref}
\usepackage{url}
\newcommand{\ourmodel}{\textsc{Weighted-PM}}
\usepackage{algorithm}
\usepackage{algorithmic}
\usepackage{graphicx}
\title{Improving Heterogeneous Graph Learning with Weighted Mixed-Curvature Product Manifold}

\author{
Tuc Nguyen-Van \\
AI Center\\
FPT Software\\
Hanoi, Vietnam \\
\texttt{tucnv2@fpt.com} \\
\And
Dung D. Le\\
VinUniversity \\
Hanoi, Vietnam \\
\texttt{dung.ld@vinuni.edu.vn} \\
\AND
The-Anh Ta \\
AI Center\\
FPT Software\\
Hanoi, Vietnam \\
\texttt{anhtt71@fpt.com}
}

\iclrfinalcopy 
\begin{document}

\maketitle

\begin{abstract}
In graph representation learning, it is important that the complex geometric structure of the input graph, e.g. hidden relations among nodes, is well captured in embedding space. 
However, standard Euclidean embedding spaces have a limited capacity in representing graphs of varying structures.
A promising candidate for the faithful embedding of data with varying structure is \emph{product manifolds} of component spaces of different geometries (spherical, hyperbolic, or Euclidean).
In this paper, we take a closer look at the structure of product manifold embedding spaces and argue that each component space in a product contributes differently to expressing structures in the input graph, hence should be weighted accordingly. 
This is different from previous works which consider the roles of different components equally.
We then propose \ourmodel, a data-driven method for learning embedding of heterogeneous graphs in \emph{weighted} product manifolds.  
Our method utilizes the topological information of the input graph to automatically determine the weight of each component in product spaces. Extensive experiments on synthetic and real-world graph datasets demonstrate that \ourmodel{} is capable of learning better graph representations with lower geometric distortion from input data, and performs better on multiple downstream tasks, such as word similarity learning, top-$k$ recommendation, and knowledge graph embedding.
We provide the source of implementation in \href{https://github.com/sharecodesubmission/weighted_product_manifold}{\textit{https://github.com/product\_manifold}}.
\end{abstract}

\section{Introduction}
Representation learning aims to acquire the ability to effectively embed meaningful data into feature spaces \cite{bengio2013representation}. In traditional representation learning models, Euclidean embedded spaces have been predominantly utilized. However, the uniform geometric structure of Euclidean spaces has certain limitations when it comes to providing accurate representations for various types of structured data, particularly graphs such as tree structures \cite{nickel2017poincare} or circular graphs \cite{wilson2014spherical}. Consequently, there is a growing interest in developing methods that enable the embedding of graph features in non-Euclidean spaces \cite{chami2019hyperbolic, topping2021understanding}.

Real-world data frequently exhibit diverse patterns and complex geometries that cannot be adequately captured by the uniform structures of Euclidean embedding spaces. It has been observed that Euclidean spaces are often insufficient for embedding various types of real-world graph data, such as hierarchical structures that induce negative curvature geometry \cite{chami2020trees}, or circle structures \cite{wilson2014spherical} that require positive curvature geometry. 
Previous research has demonstrated that using spherical embedding spaces instead of Euclidean ones can result in minimal distortion errors when embedding data with circle and ring structures \cite{wilson2014spherical,meng2019spherical}. Moreover, models that solely utilize embedding spaces of a single geometric type often struggle to capture mixed structures effectively. These models tend to produce embedding representations with significant geometric distortion compared to the underlying geometry of the input data \cite{gu2018learning}. In contrast, approaches employing product spaces composed of components with different geometries have shown promising results in graph representation learning.

\paragraph{\textbf{Problem}}
Current geometric embedding models, as seen in \cite{gu2018learning,skopek2019mixed, zhang2021switch}, typically employ product spaces with equally weighted components. In this setup, the learnable parameters are fitted to the training data samples across all component spaces in a uniform manner. However, we contend that this approach hinders the robustness of models when learning data with diverse geometric structures.
Specifically, when the input data predominantly exhibit a particular geometric type compared to others, updating all components equally may not be optimal. Instead, it would be advantageous to assign more emphasis to the dominant geometric type during the parameter update process. This would allow the model to better capture and represent the most prevalent geometric structure in the data.

\paragraph{\textbf{Our approach}}
To address this issue, we introduce a novel data-driven approach that incorporates a scoring mechanism for each component in the product spaces. This scoring mechanism enables the automatic learning of weights for each component based on the geometric structures present in the input data.
By considering the specific geometric characteristics of the data, our method allows for the construction of flexible and adaptive product spaces. This includes not only updating the weights of the components but also adjusting the geometric curvatures of the spaces.
As a result, our models are capable of effectively capturing and representing the complex geometric structures inherent in the data, leading to improved embedding performance.

\paragraph{\textbf{Contributions}} 
We summarize our contribution as follows. 
Firstly, to the best of our knowledge, this is the first work that considers the structure at each component of product manifold and proposes that each component space contributes differently to expressing various geometric structures in the input graph, hence should be weighted accordingly. 
Secondly, we propose \ourmodel{}, a data-driven method for learning to embed of
heterogeneous graphs in \textit{weighted} product manifolds.
Thirdly, we conduct extensive experiments on both synthetic and real-world datasets to validate our approach to the various downstream tasks. 

\section{Related works \& Background}
The field of machine learning has witnessed a proliferation of works focusing on learning data representations in non-Euclidean spaces, as evidenced by studies such as \cite{liu2019hyperbolic, chami2019hyperbolic, vinh2020hyperml}. However, recent research by \cite{yu2023random} has highlighted the computational challenges and numerical instability faced by hyperbolic graph convolution networks, particularly in high-dimensional settings. To address this issue, \cite{yu2023random} proposed a random feature mapping technique that utilizes the eigenfunctions of the Laplace operator to approximate an isometry-invariant kernel on hyperbolic space.
Another notable approach in this area is CurvGAN \cite{curvaGAN}, which introduces a GAN-based graph representation method that preserves the topological properties of discrete structures by approximating them as continuous Riemannian geometric manifolds. However, these methods primarily focus on a single embedding space and may struggle to effectively capture the underlying structure of the input data.

In contrast, the product of spaces has been shown to possess the capability to achieve higher generalization and effectively capture the intrinsic structures of graphs with mixed geometries \cite{gu2018learning}. By combining multiple spaces with different geometric characteristics, the product of spaces approach offers improved representation learning and a more comprehensive understanding of complex data structures.
While several approaches have explored the use of product spaces, few have addressed the challenges associated with defining and updating the component spaces. One such work, Switch Spaces \cite{zhang2021switch}, introduces a method that selects a combination of $K$ components from a set of $N$ spaces based on input specifications. It employs a gating mechanism to score and choose subspace components using pairwise relationships in the training data. However, since entities in a graph are not independent and identically distributed (iid), the component spaces selected based on individual input instances may not effectively capture the overall relationships between nodes in the graph. Consequently, Switch Spaces requires embedding spaces with high dimensions (e.g., 100, 500) to achieve competitive performance in various downstream tasks like knowledge graph embedding and recommendation.
Unfortunately, this approach unintentionally sacrifices the advantages offered by non-Euclidean models, which can achieve compactness by requiring smaller dimensions to achieve the same capacity as Euclidean space. In our study, we propose a novel approach that leverages a richer and more robust representation space to capture the diverse geometric structures present in graph data. By enhancing the quality of embeddings, our research complements existing graph-based learning methods and enables more effective representation learning.

\paragraph{\textbf{Non-Euclidean embedding spaces}}
Non-Euclidean representation learning has emerged as a powerful approach, delivering state-of-the-art performance across diverse tasks. Specifically, hyperbolic space has proven effective in tasks such as network embedding \cite{xu2022hyperminer}, recommendation systems \cite{vinh2020hyperml, you2022multi}, and knowledge graphs \cite{chami2020low, cao2022geometry}. On the other hand, spherical space excels in modeling directional similarity and data with cyclical structures \cite{meng2019spherical,gu2018learning}. Each of these spaces possesses unique geometric features, and the selection of an appropriate embedding space should be guided by the inherent structure of the data. By leveraging the most suitable embedding space, we can effectively capture the intrinsic properties and relationships within the data, leading to superior performance across a wide range of applications.

\paragraph{\textbf{Product manifold}}
Product manifolds are constructed by combining embedding spaces with different geometric types, such as Euclidean, hyperbolic, and spherical spaces. In the context of representation learning, the concept of product spaces was introduced in \cite{gu2018learning}, where each component of the product space has a constant curvature. The curvature of the product space is determined by the sum of curvatures of its individual components \cite{di2022heterogeneous}, resulting in a constant curvature overall. This property enables product spaces to capture a wide range of curvatures with lower distortion compared to a single space \cite{gu2018learning, takeuchi2022neural}. As a result, product spaces are particularly well-suited for real-world data that exhibit mixtures of geometric structures.
For example, \cite{skopek2019mixed} developed a Mixed-curvature Variational Autoencoder, which efficiently trains a VAE with a latent space consisting of a product of constant curvature Riemannian manifolds. Additionally, the heterogeneous structure present in user-item interaction graphs can be effectively learned by utilizing product spaces with different curvature components \cite{xu2022amcad}.

\paragraph{\textbf{Distortion error of embedding}} 
Given metric spaces $U$ and $V$ equipped with distances $d_U$ and $d_V$ respectively, an embedding is a continuous and injective mapping $f: U \rightarrow V$. To evaluate the quality of an embedding, we use the average distortion metric $D_\text{avg}(f)$, which calculates the average distortion over all pairs of points. Distortion between a pair of points $a$ and $b$ is defined as $\left\lvert \left(\frac{d_V(f(a), f(b))}{d_U(a, b)}\right)^2 - 1\right\rvert$. 

\section{Proposed method: \ourmodel{}}
\label{sec:proposed-model}
In this section, we present our approach to learning the weights between sub-geometries with different curvatures in the product of embedding spaces. Our objective is to ensure that the curvatures of the graph embedding spaces closely match the curvatures of the graph itself. To accomplish this, we introduce a novel gating mechanism that assigns a score to each component space.
Motivated from the coarsening approaches \cite{ying2018hierarchical,cai2021graph}, we designed gating mechanism to leverage the message-passing of information across various regions of the input graph, enabling the extraction of topology information. Our gating mechanism divides the graph into multiple parts, where each sub-graph is predominantly characterized by a specific type of geometry, such as a tree or cycle structure.
For example, in a graph consisting of a ring of trees where the tree structure dominates, we assign higher scores to hyperbolic components in the product space compared to spherical components. This choice is made to improve the quality of the embeddings produced.
By applying this gating mechanism and adjusting the weights between the different sub-geometries, we aim to achieve a more accurate representation of the graph's underlying structures, resulting in improved embedding results.

\begin{figure*}[ht]
\centering
\includegraphics[width=\textwidth]{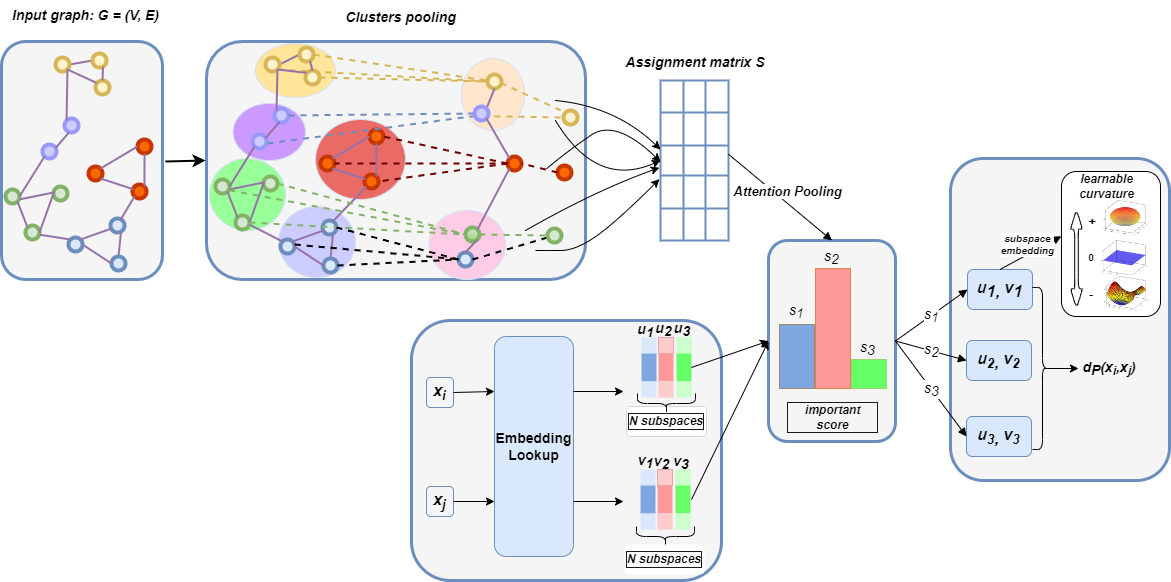}
\caption{From left to right: for an input graph $G= (V, E)$, we use a hierarchical pooling method to pool $G$ to produce assignment matrices $S_{n_l \times n_{l+1}}$, where $n_{l+1} = N$ is the number of components in the product space; self multi-head attention is then used to capture the weight vector $\mathbf{s} \in \mathbb{R}^N$ which calculates influence scores among curvatures of subspaces over all the nodes of the input graph $G$.}
\label{fig:model_architecture}
\end{figure*}

\paragraph{\textbf{Problem formulation}}
Given three types of geometry: Euclidean  ($\mathbb{E}$), Hyperbolic ($\mathbb{H}$), and Spherical ($\mathbb{S}$).
Let $\mathcal{M}_1, \mathcal{M}_2, \dots, \mathcal{M}_N$ be $N$ component spaces where $M_i$ is of one geometric type among $\{\mathbb{E}, \mathbb{H}, \mathbb{S}\}$, and $\dim M_i = b_i$.
The goal of our approach is to learn the score $\mathbf{w} = (w_1, \dots, w_N) \in \mathbb{R}^N$ from the input graph data on each component of product manifold embedding space in such a way that the embedding of input graph into $P = w_1 \mathcal{M}_1 \times w_2 \mathcal{M}_2 \times \dots \times w_N \mathcal{M}_N$ will have lowest possible geometric distortion.

\subsection{Coarsening input graph data}
\label{coarsening_module}
\paragraph{\textbf{Hierarchical pooling layers}}
Given input graph $\mathcal{G}$, with $n > 0$ nodes, adjacency matrix $\mathbf{A} \in \{ 0, 1\}^{n \times n}$ and node features $\mathbf{X} \in \mathbf{R}^{n \times d}$.
The matrix $\mathbf{A}$ represents graph structure: $\mathbf{A}(i, j) = 1$  if there is an edge connecting two nodes $i, j$, otherwise $\mathbf{A}(i, j) = 0$.
$D$ is the diagonal degree matrix of the graph $\mathbf{G}$ where $D_{ii} = \sum_i \mathbf{A}_{ij}$.

We use hierarchical pooling-based GCNs to learn cluster assignments. 
There are two GCNs with two different sets of parameters in this module. 
At each layer $l$, the soft cluster assignment matrix $\mathbf{S}^{(l)} \in \mathbf{R}^{n_{l-1} \times n_l}$ is computed as follows:
\begin{equation}
\resizebox{0.8\hsize}{!}{
    $\mathbf{S}^{(l)} = \operatorname{softmax} (GNN_1^l(\mathbf{A}^{(l-1)}, \mathbf{X}^{(l-1)}))$ \text{ with } $(\mathbf{A}^{(0)}, \mathbf{X}^{(0)}) =  (\mathbf{A}, \mathbf{X})$}.
\end{equation}
Then, we apply the second GNN on $\mathbf{S}^{(l)}$ to compute the graph representation at layer $l$:
\begin{equation}
\resizebox{0.8\hsize}{!}{
    $\mathbf{X}^{(l)} = {\mathbf{S}^{(l)}}^T (GNN_2^{(l)}(\mathbf{A}^{(l-1)}, \mathbf{X}^{(l-1)}))$ \text{ and } $\mathbf{A}^{(l)} = {\mathbf{S}^{(l)}}^T \mathbf{A}^{(l-1)} \mathbf{S}^{(l)})$}.
\end{equation}

\paragraph{\textbf{Coarsening input graph}}
The hierarchical pooling layer produces a coarsened graph with $m < n$ nodes, a weighted adjacency matrix $A' \in \mathbb{R}^{m \times m}$, and node embeddings $Z' \in \mathbb{R}^{m \times d}$. 
This process is then repeated $L$ times, resulting in a GNN model with $L$ layers that operate on the input graph and a series of coarser versions of it. 
The soft assignment matrix $S^{(l)}$ assigns each node at layer $l$ to a cluster at the next layer $l+1$. 
In other words, each row of $S^{(l)}$ corresponds to one of the $n_l$ nodes or clusters at layer $l$, while each column of $S^{(l)}$ corresponds to one of the $n_{l+1}$ clusters at layer $l+1$. 
In our approach, we treat the number of clusters as a hyperparameter and set $n_{l+1} = N$, where $N$ is the number of components in the product space $P$. 
Each row of $S^{(l)}$ shows the degree of membership of a node to each component space in $P$.

\paragraph{\textbf{Attention pooling}}
We use the attention mechanism with the input being the matrix $\mathbf{S}^{(l)}$ to take the influence vector for each subspace.
Consider the matrix $\mathbf{S}$ in form $\mathbf{S} = [\mathbf{h}_1, \mathbf{h}_2, ... , \mathbf{h}_N ]$, with $\mathbf{h}_t \in \mathbb{R}^d$ , and a trainable matrix $\mathbf{U} \in \mathbb{R}^d$.

\textit{Self attention:}
We define a relevant scalar weight for each element of the sequence through a softmax layer as follows $w_t = \operatorname{softmax}(\mathbf{h}_t^T \mathbf{U})$.
Given the set of weights over all the elements of the sequence, we can then obtain the pooled representation as the weighted average of the hidden states
$\textbf{s} = \sum_{t = 1}^N \mathbf{h}_t^T \mathbf{w}_t$.

\textit{Multi-head self attention:}
Considering a number of $k$ heads for the multi-head attention, $\mathbf{h}_t = [\mathbf{h}_{t1},  \mathbf{h}_{t2}, \dots, \mathbf{h}_{tk}]$ where $\mathbf{h}_{tj} \in \mathbb{R}^{d/k}$ and size of each head is $d/k$.
%
%
In the same sense, we have a trainable parameter $\mathbf{U} =[\mathbf{u}_1 \mathbf{u}_2 \dots \mathbf{u}_k]$ where $\mathbf{u}_j \in \mathbb{R}^{d/k}$.
Different attention is then applied over each head of the encoded sequence softmax function following 
$w_{t} = \operatorname{softmax}(\mathbf{h}_{tj}^T \mathbf{u}_j)$, where $w_{tj}$ corresponds to the attention weight of the head $j$ on the element $t$. 
A soft weight representation for each subspace is computed as follows:
\begin{equation}
    s_j = \sum_{t=1}^N \mathbf{h}_{tj}^T \mathbf{w}_{tj}.
\end{equation}
This method allows a multi-head self-attention network extracts different kinds of information over different regions of the self-attention network.
In the end, $\mathbf{s} \in \mathbb{R}^N$ represents the average weight of $N$ component spaces in the product manifold $P$ over the $n_l$ clusters.

\begin{algorithm}[tb]
\caption{R-SGD in the product of model spaces}
\label{alg:algorithm}
\begin{algorithmic}[1]
\STATE Initialize $x^{(0)} \in \mathcal{P}$
\FOR{t = 0, \dots, T - 1}
\STATE $h \leftarrow \Delta \mathcal{L}_{distortion}(x^{(t)})$

\STATE \textit{{Step:} Project positive curvature embedding to the tangent space.}
\FOR {i = 1, \dots, m}
\STATE $v_j \leftarrow \operatorname{proj}_{x_i^{(t)}}^{S} (h_i)$
\ENDFOR
\STATE \textit{{Step:} Project negative curvature embedding to the tangent space.}
\FOR {i = m + 1, \dots, m + n}
\STATE $v_i \leftarrow \operatorname{proj}_{x_i^{(t)}}^{H} (h_i)$
\STATE $v_i \leftarrow Jv_i$
\ENDFOR
\STATE \textit{{Step:} Euclidean space.}
\STATE $v_{m+n+1} \leftarrow h_{m+n+1}$
\FOR {i = 1, \dots, m}
\STATE $v_j \leftarrow \operatorname{proj}_{x_i^{(t)}}^{S} (h_i)$
\ENDFOR
\STATE \textit{{Step:} Curvature-wise concatenation.}
\STATE $v_{m+n+1} \leftarrow h_{m+n+1}$
\STATE \textit{{Step:} Project from tangent space to the manifold.}
\FOR{i = 1, \dots, m +n + 1}
\STATE $x_{i}^{(t+1)} \leftarrow \operatorname{Exp}_{x_i^{(t)}} (v_i)$
\ENDFOR 
\ENDFOR
\STATE $\textbf{return } x^{(T)}$ \\
\end{algorithmic}
\end{algorithm}

\subsection{Objective function}
Let $\mathbf{s} \in \mathbb{R}^N$ be the weight vector of $N$ components based on the data's local geometry information. 
The distance between $x_i, x_j \in P$ is computed following $d_P^2(x_i, x_j) = \sum\limits_{k = 1}^N \mathbf{s}_k \operatorname{dist}^2 (x_i^k, x_j^k)$.
Then the base objective $\mathcal{L}_{base}$ is defined as:
\begin{equation}
    \mathcal{L}_{base}   = \sum_{1 \leq i < j \leq n}\left|\left(\frac{d_{P}\left(x_{i}, x_{j}\right)}{d_{G}\left(X_{i}, X_{j}\right)}\right)^{2}-1\right|
\end{equation}

Finally, the total average distortion objective function is defined as $\mathcal{L} = \mathcal{L}_{base} + \mathcal{L}_{aux}$, 
where $\mathcal{L}_\text{aux} = \mathcal{L}_\text{LP} + \mathcal{L}_e$ is a combination of the link prediction loss ($\mathcal{L}_\text{LP}$) and the entropy regularization loss ($\mathcal{L}_e$). 
More precisely, $\mathcal{L}_\text{LP} = \left\|\mathbf{A}^{(l)} - \mathbf{S}^{(l)} \mathbf{S}^{(l)^{T}}\right\|_{F}$ at each layer $l$, where $\|\cdot\|_{F}$ denotes the Frobenius norm; and
$\mathcal{L}_e=\frac{1}{n} \sum\limits_{i=1}^{n} H\left(\mathbf{S}_{i}\right)$ where $H(\mathbf{S}_i)$ is the entropy of the row $i^{th}$ in matrix $\mathbf{S}$.
Minimizing  $\mathcal{L}_\text{LP}$ means enforcing close nodes to be pooled together, while 
minimizing $\mathcal{L}_e$ makes the output cluster assignment for each node close to a one-hot vector so that the membership for each cluster is clearly defined.
Our total average distortion $\mathcal{L}$ is optimized with the Algorithm \ref{alg:algorithm}.

\subsection{Physical meaning of subspace weights}
In manifold representation learning, the goal is to embed data into appropriate embedding spaces where the curvature of the embedding matches the curvature of the original data. In the case of a product manifold, each data point is partially embedded in different subspaces with varying curvatures.
Our work explores the relationship among the curvatures of all the subspaces and introduces a partial update mechanism for the embedding space based on their respective influence scores. In the importance score box of Model Architecture (Figure \ref{fig:model_architecture}), if the input data is predominantly characterized by hierarchical structures, the importance score of the hyperbolic embedding component ($s_2$) will receive a larger value compared to the others ($s_1$ and $s_3$).
In Algorithm \ref{alg:algorithm}, we update the subspaces' curvatures and the embedding itself. The higher the curvature embedding scores, the more effort is required to minimize them. As a result, the negative curvature loss should contribute more to the overall loss, leading to more active updates of the embedding spaces associated with negative curvature compared to the other spaces. This ensures that the embedding adapts to the data's curvature characteristics and effectively captures the underlying structures.

\section{Experiments}
\label{section:experiments}
This section presents our experimental evaluation of the proposed model's performance across various learning tasks. We begin by evaluating the model's effectiveness in improving graph reconstruction, as described in section \ref{graph_reconstruction}.
Following this, we apply our framework to four downstream tasks: recommendation systems, knowledge graph embedding, node classification, as well as graph classification, and word similarity tasks.

\subsection{Graph reconstruction}
\label{graph_reconstruction}

We perform experiments on both synthetic and real-world datasets to evaluate the performance of our proposed model. 
More information on baselines and metrics is shown in Appendix \ref{apendix_graph_reconstruction}.

\paragraph{\textbf{Model performance on synthetic datasets}}
Table \ref{tab:synthetic_result} shows the average distortion ($D_\text{avg}$) of our model on the three synthetic graphs. When $d = 3$, \ourmodel{} achieves $D_{avg} = 0.104$ with the product manifold $\textbf{s}_1 \mathbb{H}^{2} \times \textbf{s}_2 \mathbb{S}^{1}$. 
Meanwhile, without any constraints in subspace curvatures (PM \cite{gu2018learning}), the distortion measure of $\mathbb{H}^{2} \times \mathbb{S}^{1}$ on the Cycle graph is $0.11$.
Overall, for all three synthetic graphs, our proposed model improves upon the main contender method PM from \cite{gu2018learning} by $5.4 \%$, $16.3 \%$, and $18.6 \%$, respectively (Table \ref{tab:synthetic_result}).
Similar trend continues in higher dimension $d=5$, our proposed method improves upon the baseline by $17.3\%$, $3.3\%$ and $11.9 \%$, respectively (Table \ref{tab:synthetic_result}).

\begin{table}[t]
\centering
\caption{$D_{avg}(\downarrow)$ results on the synthetic data sets with the dimension $d$ equal to 3 and 5 respectively. $\mathbf{s}_i$ are learnable weights depending on input data.}
\vspace{1mm}
\resizebox{0.6\columnwidth}{!}{%
\begin{tabular}{c c c c c}
\hline

{Method}  &{Models} 
 &{Cycle}  &{Tree}  &{Mix}\\ 
\hline 
{PM}                  &          \\
&$\mathbb{H}^{2} \times \mathbb{S}^{1}$          &      0.110        &    0.055 & $\mathbf{0.063}$  \\

{\ourmodel{}}                  &          \\
&$ \textbf{s}_1 \mathbb{H}^{2} \times \textbf{s}_2 \mathbb{S}^{1}$          &      $\mathbf{0.104}$       &    0.051 & 0.058  \\ 
&         &     \tiny{(+5.4\%)}       &     &  \\ 

&$ \textbf{s}_1 \mathbb{H}^{1} \times \textbf{s}_2 \mathbb{S}^{2}$          &      0.116        &    $\mathbf{0.046}$ & $\mathbf{0.051}$  \\ 

&          &          &    \tiny{(+16.3\%)} & \tiny{(+18.6\%)}  \\ 
\hline

{PM}                  &          \\
&$ \mathbb{H}^{3} \times \mathbb{S}^{2}$          &   0.124     &  $\mathbf{0.06}$   &  0.054 \\ 
&$ \mathbb{H}^{2} \times \mathbb{S}^{3}$          &   $\mathbf{0.115}$     &  0.07   & 0.052  \\ 
&$ \mathbb{H}^{2} \times \mathbb{S}^{2} \times \mathbb{E}$          &    0.12    &  0.092   & $\mathbf{0.047}$  \\

{\ourmodel}                  &          \\
&$\textbf{s}_1 \mathbb{H}^{3} \times \textbf{s}_2 \mathbb{S}^{2}$          &    0.102    &  0.068   &  0.05 \\ 
&$ \textbf{s}_1 \mathbb{H}^{2} \times \textbf{s}_2 \mathbb{S}^{3}$          &     0.11   &  $\mathbf{0.058}$   &  0.054 \\ 
&          &      &  \tiny{(+3.3\%)}   &  \\ 

&$\textbf{s}_1 \mathbb{H}^{2} \times \textbf{s}_2\mathbb{S}^{2} \times \textbf{s}_3\mathbb{E}$          &   $\mathbf{0.095}$     &  0.086   &  $\mathbf{0.041}$ \\
&         &    \tiny{(+17.3\%)}    &    &  \tiny{(+11.9\%)} \\
\hline
\end{tabular}}
\label{tab:synthetic_result}
\end{table}

\begin{figure}[!h]
    \centering
    \includegraphics[scale = 0.5]{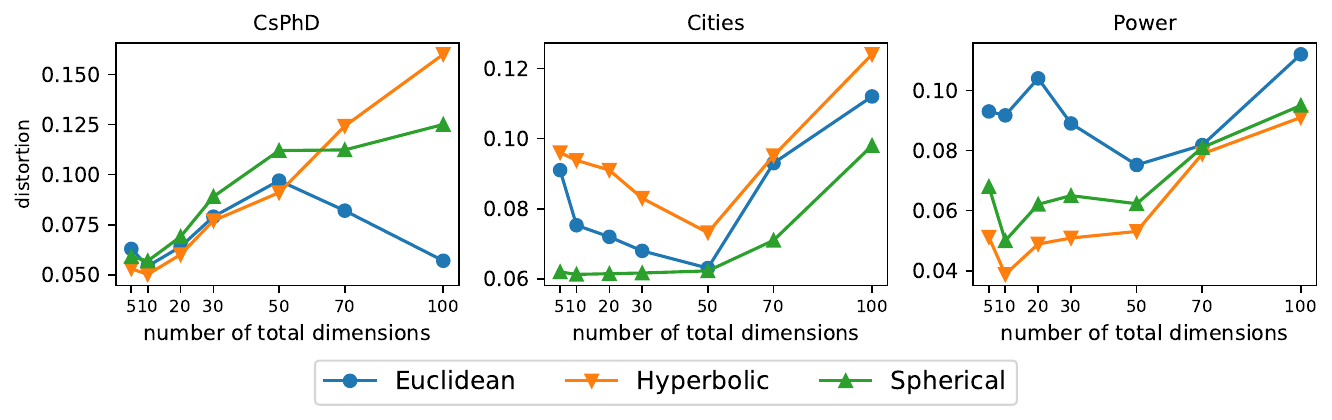}
   \caption{Embedding benchmark datasets on one single space with the number of dimensions increasing gradually.}
   \label{distortion_csphd_power_cities}
\end{figure}

\paragraph{\textbf{Model performance on benchmark datasets}}
We first employ a single space to generate embedding representations for each dataset in order to explore its intrinsic geometry. 
Based on these observations, we develop heuristics for the dataset characteristics and utilize them to select the component in the model space product. 
Then, the learning process optimizes the curvature of each subspace according to the dominant graph structure. 
Figure \ref{distortion_csphd_power_cities} presents the average distortion $D_\text{avg}$ of embeddings into single model spaces for three complex benchmark datasets, as the number of embedding dimensions increases within the range of $[5, 100]$. 
We can see that, with the Cs PhDs and Power dataset, $D_\text{avg}$ is smaller in hyperbolic space than in spherical space when $d<50$, indicating that the hyperbolic space should be considered in the general product space. 
Similarly, the Cities dataset exhibits a more spherical structure than other geometric properties, and thus components of positive curvature should be used.

Table \ref{tab:benchmark_csPhD_datasets} reports the performance of our model on the benchmark datasets. 
Unlike the results obtained from the synthetic dataset, the best results are predominantly obtained when learning with the product manifolds. 
This phenomenon is attributed to the more complex structure of real-world data compared to synthetic ones. 
Specifically, the Power graph dataset has a hierarchical and cyclical structure that can be embedded effectively into any space with a hyperbolic and spherical component. 
Our proposed model outperforms the main baseline PM \cite{gu2018learning} in all cases. 
With embedding dimension $d = 10$, our model achieves the best distortion on the three datasets. 
Specifically, in the Cs PhDs dataset, the percentage of improvements in terms of $D_\text{avg}$ is $15.6 \%$. 
In the Power dataset, with the soft gating mechanism, our model achieves better distortion upon the product of the space model with $28.4 \%$. 
In case $d = 50$, the same improvement with $d = 50$, for specific, these average distortions ($D_\text{avg}$) compare with the uniform product of spaces (PM) of \cite{gu2018learning} is $19.3 \%$ and $13.9\%$, respectively.

Furthermore, Table \ref{tab:benchmark_csPhD_datasets} shows that for distortion of $0.0231$ in the product space $\mathbb{H}^5 \times \mathbb{S}^5$ with the Power dataset, our method determines that the optimally weighted product manifold for embedding the dataset are $0.83 \,\mathbb{H}^5 \times 0.16 \, \mathbb{S}^5$. The ratio between the hyperbolic and spherical components is approximately $5:1$, indicating the greater importance of hyperbolic components compared to spherical ones. 
In contrast, the uniform product embedding space PM of \cite{gu2018learning} assumes that each component space contributes equally to learning representations in the product of spaces. 

Our method \ourmodel{}, on the other hand, captures the constraints relation among all sub-geometries of different curvatures in the product manifold, depending on the geometry of the input graph data, leading to better performance than using the uniform product of spaces (PM) without scoring mechanism. Our proposed method has advantages in discovering general models with suitable geometry in the product manifold. Notably, we also observe that the $mAP$ measures are not consistently better than the uniform product model spaces \cite{gu2018learning} when $D_\text{avg}$ decreases.

\begin{table}[ht]
\centering
\caption{Average distortion $D_\text{avg}(\downarrow)$ and mAP $(\uparrow)$ on CsPhDs and Power datasets on dimensions 10 and 50 respectively. The best $D_\text{avg}$ results are in boldface, while the second best results are underlined.}
\vspace{1mm}
\resizebox{0.8\columnwidth}{!}{%
\begin{tabular}{c c c c c c c}
\hline

\multicolumn{1}{c}{Method} &\multicolumn{1}{c}{Models} &  \multicolumn{2}{c}{Cs PhDs} &  \multicolumn{2}{c}{Power} \\ 
 && \multicolumn{1}{c}{$D_\text{avg}$} &\multicolumn{1}{c}{mAP} & \multicolumn{1}{c}{$D_\text{avg}$} &\multicolumn{1}{c}{mAP}\\

\hline
\small{PM} &$(\mathbb{H}^{2})^5$           & $\underline{0.0357}$ & 0.9694   & 0.0396 & 0.8739\\ 
&$\mathbb{H}^{5} \times \mathbb{S}^{5}$          & 0.0529 & 0.9041  & $\underline{0.0323}$ & 0.8850 \\ 

\small{\ourmodel}                  &          \\
&$\mathbf{s}(\mathbb{H}^{2})^5$          & $\mathbf{0.0301}$ & 0.9699  & 0.0423 & 0.8540 \\ 
&         & \tiny{($+15.6\%$)}   \\ 

&$\mathbf{s}_1\mathbb{H}^{5} \times \mathbf{s}_2\mathbb{S}^{5}$           & 0.0494 & 0.9231 & $\mathbf{0.0231}$ & 0.8842 \\ 
&& &           & \tiny{($+28.4\%$)}    \\
\hline
\small{PM}                  &          \\
&$(\mathbb{H}^{5})^{10}$          & 0.0786 & 0.8600   & 0.0723 & 0.842 \\ 
&$(\mathbb{H}^{10})^3 \times (\mathbb{S}^{10})^2 $          & 0.0620 & 0.9120  & 0.0510 & 0.8500 \\ 
&$(\mathbb{H}^{10})^2 \times (\mathbb{S}^{10})^2 \times \mathbb{E}^{10}$          & $\underline{0.0571}$ & 0.9071  & $\underline{0.0495}$ & 0.8561  \\ 

\small{\ourmodel}                  &          \\
&$\mathbf{s}(\mathbb{H}^{5})^{10}$          & 0.0591 & 0.8900  & 0.0503 & 0.8570\\ 

&$\mathbf{s_1}(\mathbb{H}^{10})^3 \times \mathbf{s_2}(\mathbb{S}^{10})^2 $          & 0.0590 & 0.9312  & $\mathbf{0.0426}$ & 0.8620 \\ 
& & &  & \tiny{($+13.9 \%$)}  \\ 

&$\mathbf{s_1}(\mathbb{H}^{10})^2 \times \mathbf{s_2}(\mathbb{S}^{10})^2 \times \textbf{s}_3 \mathbb{E}^{10}$          & $\mathbf{0.0460}$ & 0.9240  & 0.0502 & 0.8600\\ 
&  & \tiny{($+19.3 \%$)}  \\

\hline
\end{tabular}}
\label{tab:benchmark_csPhD_datasets}
\end{table}

\subsection{\ourmodel{} on Knowledge Graph Embedding}
Knowledge graphs (KGs) are a fundamental tool for representing information and have a wide range of applications, including question answering and web search \cite{ji2021survey}. 
However, KGs are often highly incomplete, which poses a significant challenge for downstream use. 
The goal of our approach is to address this issue by inferring missing facts in the KGs using entity and relation embedding techniques to map them to appropriate spaces. 
In this section, we propose using the product of manifolds with a gating mechanism to represent the relations between entities in the KGs.
Detailed experimental scenario is shown in Appendix \ref{apendix_kg_embedding}.
\paragraph{\textbf{Model performance}}
Table \ref{tab:knowledge_graph} reports the performance of various methods on two knowledge graphs. 
To enable a fair comparison, we set the total embedding dimension to 64, which is a common practice in non-Euclidean embedding due to its ability to provide more compact spaces than Euclidean embeddings. 
Our proposed model achieves superior performance over the baselines on the knowledge embedding graph, highlighting its effectiveness in learning informative representations of the data.

\begin{table}[ht]
\centering
\caption{Model performance on knowledge graph embedding task. The best results are in boldface, while the second-best results are underlined.}
\vspace{1mm}
\resizebox{0.7\columnwidth}{!}{%
\begin{tabular}{c c c c c}
\hline
\multicolumn{1}{c}{Best model} &  \multicolumn{2}{c}{FB15k-237}  &  \multicolumn{2}{c}{WN18RR}\\ 
 
& {MRR ($\uparrow$)} & {HR@3 ($\uparrow$)} & {MRR ($\uparrow$)} & {HR@3 ($\uparrow$)} \\

\hline
RotatE ($E^{64}$)  & 0.228  &  0.265 & 0.216 &   0.384  \\
RotatH ($H^{64}$) & 0.292 & 0.320 &  0.340 & 0.421    \\
Product-RotatH ($(H^{16})^4$) & 0.321 & 0.332 &  0.352 & 0.451 \\
SwisE & \underline{0.334} & \underline{0.337} &  \underline{0.378} & \underline{0.462}     \\
    \ourmodel-Rotat & \textbf{0.339} & \textbf{0.345} &  \textbf{0.392} & \textbf{0.469}     \\
\hline

\end{tabular}}
\label{tab:knowledge_graph}
\end{table}

\subsection{\textbf{\ourmodel{} on node classification and link prediction}}
In this section, we evaluate the performance of our proposed model on node and graph classification tasks. 
Hyperbolic GCN \cite{chami2019hyperbolic} uses message-passing on the hyperbolic tangent space for graph convolutional networks (GCNs). 
However, our proposed model replaces the hyperbolic space with \ourmodel{} and applies message passing in the tangent of the product spaces. 
We further introduce $\delta$ \cite{bridson2013metric} which is used to evaluate the degree of tree-likeness of a graph by evaluating its graph distance metric.
The value of $\delta$ ranges from 0 to half of the graph diameter, with trees having $\delta = 0$, while "circle graphs" and "grid graphs" have a larger $\delta$, approximately half of their diameters.
Further details on the metrics, datasets, and baselines used in our experiments can be found in Appendix \ref{apendix_node_graph_cls}.

\paragraph{\textbf{Model performance}}
Table \ref{tab:disease_task} presents the F1 and AUC scores for the link prediction and node classification tasks. 
Notably, the DISEASE and AIRPORT datasets exhibit high hyperbolicity ($\delta$ = 0 and 1, respectively), where the performance of using the product of hyperbolic space surpasses that of using the product of mixture curvatures. 
This is because the unified product of curvature fails to differentiate the primary intrinsic graph structure and instead adapts equally to spaces that do not align with the graph's topology. 
Our proposed extension addresses this issue by incorporating a weighting mechanism that identifies the dominant embedding manifold most influenced by the underlying structure of the graph data, leading to improved results in both link prediction and node classification for these two datasets.

\begin{table}[ht]
\centering
\caption{F1 ($\uparrow$) and AUC ($\uparrow$) scores on DISEASE and AIRPORT datasets. The best results are in boldface, while the second-best results are underlined.}
\vspace{1mm}
\resizebox{0.85\columnwidth}{!}{%
\begin{tabular}{c c c c c}
\hline
\multicolumn{1}{c}{Model}  & \multicolumn{2}{c}{DISEASE} & \multicolumn{2}{c}{AIRPORT} \\ 
 
& {F1} & {AUC} & {F1} & {AUC} \\

\hline
MLP (${E}^{16}$) &  28.8 $\pm$  2.5  & 72.6 $\pm$  0.6 &  68.6 $\pm$  0.6 &   89.8 $\pm$  0.5 \\
MLP (${H}^{16}$) & 41.0 $\pm$ 1.8 & 75.1 $\pm$ 0.3  & 80.5 $\pm$ 0.5 &  90.8 $\pm$ 0.2  \\
HGCN \cite{chami2019hyperbolic} (${H}^{16}$) & 74.5 $\pm$ 0.9 & 90.8 $\pm$ 0.3   & 90.6 $\pm$ 0.2  & \underline{96.4 $\pm$ 0.1}  \\
Product-HGCN \cite{takeuchi2022neural} (${H}^{8}\times{H}^{8}$) & \underline{74.9 $\pm$ 0.6} & \underline{91.1 $\pm$ 0.2}    & \underline{90.5 $\pm$ 0.1}  & 95.7 $\pm$ 0.2 \\
Mix-GCN \cite{gu2018learning} (${H}^{5}\times{S}^{5}\times{E}^{6}$) & 74.1 $\pm$ 0.3 & 90.8 $\pm$ 0.1   & 90.1 $\pm$ 0.2  & 95.3 $\pm$ 0.1 \\
Ours (${H}^{5}\times{S}^{5}\times{E}^{6}$) & \textbf{75.4 $\pm$ 0.4} & \textbf{93.6 $\pm$ 0.2} & \textbf{93.0 $\pm$ 0.1}  & \textbf{97.6 $\pm$ 0.2} \\
\hline
\medskip
\end{tabular}}
\label{tab:disease_task}
\end{table}

 


\subsection{\ourmodel{} on Recommendation Systems}
In this section, we evaluate the performance of our proposed model on the recommendation task. Specifically, we apply \ourmodel{} to replace the hyperbolic space in metric learning recommendation (HyperML \cite{vinh2020hyperml}). Detailed information on baselines, datasets and metrics can be seen in Appendix \ref{apendix_rec_sys}.

\paragraph{\textbf{Objective function}}
In HyperML \cite{vinh2020hyperml}, the push-pull loss is proposed to learn the metric between the positive and negative items. 
The overall objective is defined as $
    \mathcal{L} = \mathcal{L}_P + \gamma \mathcal{L}_D,$
where pull-push loss $\mathcal{L}_P$ and distortion loss $\mathcal{L}_D$ are defined as: 
\begin{equation}
    \mathcal{L}_P = \sum_{(i, j) \in \mathbb{S}} \sum_{(i, k) \not\in \mathbb{S}} [m + d^2_\mathbb{D}(i,j) - d^2_\mathbb{D}(i,k)]_+,
\end{equation}
\begin{equation}
\resizebox{0.8\hsize}{!}{
$\mathcal{L}_D = \sum\limits_{(i, j) \in \mathbb{S}} \Big[\frac{d_\mathbb{D}(f(i), f(j)) - d_\mathbb{E}(i, j)|}{d_\mathbb{E}(i, j)} \Big]_+ + \sum\limits_{(i, k) \not\in \mathbb{S}} \Big[\frac{d_\mathbb{D}(f(i), f(k)) - d_\mathbb{E}(i, k)}{d_\mathbb{E}(i, k)} \Big]_+,$
}
\end{equation}
where $|z|_+ = \operatorname{max}(0, z)$, $m > 0$ is the margin size ($m = 0.4$ in this paper), 
and $f(.)$ is a mapping function $f: \mathbb{E} \rightarrow \mathbb{D}$ ($f$ is the identity in \cite{vinh2020hyperml}), $\gamma$ is the multi-task learning weight and $\mathbb{S}$ is the set of positive user-item pairs. 
We use the same loss function in \cite{vinh2020hyperml} with a difference in the distance on $\mathbb{D}$. For specific, we compute the distance $d$ between two embeddings in the product of model spaces.

\paragraph{\textbf{Model performance}}
\begin{table}[!htb]
    \begin{minipage}[l]{.48\linewidth}
    \caption{\small $H@10 (\uparrow)$ and $N@10(\uparrow)$ results of on the MovieLens 1M and LastFM-20K datasets with dimension $d$ are 32 and 64 respectively.}
    \label{tab:recsys_task}
      \centering
        \resizebox{0.9\columnwidth}{!}{%
        \begin{tabular}{c c c c c c}
            \hline
            \multicolumn{1}{c}{Method}  &\multicolumn{1}{c}{Model} &  \multicolumn{2}{c}{MovieLens}  &  \multicolumn{2}{c}{LastFM}\\ 
             
            && {H@10} & N@10 & {H@10} & N@10 \\
            
            \hline
            {CML}   & $\mathbb{E}^{32}$ & 0.641  & 0.420  & 0.171 & 0.113    \\
            {HyperML}& $\mathbb{H}^{32}$ & 0.655 & 0.435  & 0.173 &  0.123   \\
            \medskip
            {HyperML-PM}   & $(\mathbb{H}^{8})^{4}$ & \underline{0.702} & \underline{0.473}  & \underline{0.190} & \underline{0.136}
            \\
            & $(\mathbb{H}^{16})^{2}$ & 0.682 & 0.465 & 0.180 & 0.129
            \\
            \medskip
            {HyperML-WPM}  &  $\mathbf{s}(\mathbb{H}^{8})^{4}$ & \textbf{0.710} & \textbf{0.483} & \textbf{0.193} & \textbf{0.140}\\ 
            &   $\mathbf{s}(\mathbb{H}^{16})^{2}$   & 0.672 & 0.460  & 0.180 & 0.129\\ 
            \hline
            {CML}   & $\mathbb{E}^{64}$ & 0.721    & 0.513  & 0.163 & 0.098 \\
            {HyperML}   &   $\mathbb{H}^{64}$  & 0.739 & 0.520  & 0.166 &  0.101   \\
            
            \medskip
            {HyperML-PM}   &  $(\mathbb{H}^{16})^{4}$     & 0.740 & 0.520 & \textbf{0.182} & $\mathbf{0.110}$\\
            &  $(\mathbb{H}^{32})^{2}$       & \underline{0.750} & \underline{0.521}  & 0.172 & 0.105\\
            \medskip
            {HyperML-WPM}  &  $\mathbf{s}(\mathbb{H}^{16})^{4}$          & 0.732 & 0.496 & \underline{0.180} & \underline{0.108} \\ 
            &  $\mathbf{s}(\mathbb{H}^{32})^{2}$ & \textbf{0.752} & \textbf{0.522}  & 0.169 & 0.101\\
            \hline
            \end{tabular}}
    \end{minipage}%
    \begin{minipage}[r]{.43\linewidth}
      \centering
        \caption{\small Spearman rank correlation ($\uparrow$) on the WS-353 dataset on the weighted product manifold with 10 and 50 dimensions, respectively.}
        \label{tab:spearman_task}
        \resizebox{0.9\columnwidth}{!}{%
        \begin{tabular}{c c c}
            \hline
            \multicolumn{1}{c}{Method}  &\multicolumn{1}{c}{Best model} 
             &\multicolumn{1}{c}{Spearman rank}\\ 
            \hline 
            {Single} & $\mathbb{H}^{10}$       & 0.4412  \\ 
            \medskip
            {PM} & $\mathbb{H}^{5} \times \mathbb{H}^{5} $          &      \underline{0.4489} \\ 
            \medskip
            {\ourmodel}  & $ \mathbf{s}_1 \mathbb{H}^{5} \times \mathbf{s}_2 \mathbb{H}^{5}$          &      \textbf{0.4510}\\ 
            \hline
            {Single} &   $\mathbb{H}^{50}$       &  0.6389 \\
            \medskip
            {PM} &$ (\mathbb{H}^{25})^2$          &   0.6421\\ 
            &$ (\mathbb{H}^{10})^5$          &    \underline{0.6531}\\ 
            \medskip
            
            {\ourmodel} &$ \mathbf{s}_1 \mathbb{H}^{25} \times \mathbf{s}_2 \mathbb{H}^{25}$          &     0.6506 \\
            &$ \mathbf{s} (\mathbb{H}^{10})^5$ &   \textbf{0.6612} \\
            \hline
            \end{tabular}}
    \end{minipage} 
\end{table}

Table \ref{tab:recsys_task} reports the $H@10$ and $N@10$ scores for two different datasets, considering the number of factors $d \in \{32, 64\}$.
Our experiments demonstrate that, overall, CML and HyperML achieve better results with the weighted product manifolds (\ourmodel) than in the Hyperbolic space alone, highlighting the advantages of using scoring sub-manifolds to model the distance between users and items.

\subsection{Performance on word similarity task}
We evaluated our model's performance on applications that require an understanding of the underlying manifold structure. To conduct our experiment, we trained word embeddings on the Word Similarity (WS-353) benchmark dataset, following the methodology established in previous works such as \cite{gu2018learning,leimeister2018skip}. Our implementation is based on hyperbolic skip-gram embeddings from \cite{leimeister2018skip}.

\paragraph{\textbf{Setup}}
For our setup, we utilized the standard skip-gram model \cite{mikolov2013efficient} and extended the loss function to a generic objective suitable for arbitrary manifolds, using a variant of the objective used in \cite{leimeister2018skip}. 
Specifically, given a word $u$ and a target $w$ with label $y=1$ if $w$ is a context word for $u$ and $y=0$ if it is a negative sample, our model is represented by $P(y \mid w, u)=\sigma\left((-1)^{1-y}\left(-\cosh \left(d\left(\alpha_{u}, \gamma_{w}\right)\right)+\theta\right)\right)$.

\paragraph{\textbf{Word similarity}}  
To measure the effectiveness of our model, we evaluated its performance on the WS-353 dataset using the Spearman rank correlation $\rho$ between our scores and annotated ratings. 
We obtained the dataset from \cite{finkelstein2001placing}, and the results of our experiment are presented in Table \ref{tab:spearman_task}. 
Our model outperformed the hyperbolic word embeddings of \cite{leimeister2018skip} and the product space (PM) in all dimension settings.

\section{Conclusions}
\label{sec:conclusion}
Real-world data often possess intricate geometric structures that are challenging to capture by embedding into spaces with uniform curvature. 
To address this issue, we propose a method that partially extracts the topology information from the input data to update the embedding vectors and curvature of each subspace. 
Our motivation is that graphs are constructed by combining simple structure topologies, such as trees, cycles, and stars. 
Our approach introduces a data-driven method of weighted product spaces for learning better representations. 
Our empirical experiments on synthetic and real-world datasets demonstrate that our framework enhances the embedding quality of input graphs with varying structures and improves the performance of the downstream tasks.

\bibliography{main}
\bibliographystyle{iclr2021_conference}

\newpage
\appendix
\section{Additional background}
\paragraph{\textbf{Riemannian Geometry}}
Let $\mathcal{M}^n$ be a smooth manifold in $n$-dimensional space, where $\mathcal{M}^n$ is locally approximated by an $n$-dimensional Euclidean tangent space $T_p\mathcal{M}$ at $p \in \mathcal{M}$. 
The pair $(\mathcal{M}, g)$ is called a Riemannian manifold if $\mathcal{M}$ is equipped with a positive-definite metric tensor $g$ that satisfies certain conditions. 
Geodesics are the shortest-distance paths on manifolds, and the metric tensor $g$ is integrated along the geodesic to compute distances on a Riemannian manifold. 
The exponential map $\operatorname{exp}_p: T_p \mathcal{M} \rightarrow \mathcal{M}$ and logarithmic maps $\operatorname{log}_p: \mathcal{M} \rightarrow T_p \mathcal{M}$ are two common bijections defined on the manifold $\mathcal{M}$. 
A formal introduction to Riemannian manifolds can be found in \cite{petersen2006riemannian}.

\paragraph{\textbf{Product manifolds}}
Consider a sequence of smooth Riemannian manifolds $\mathcal{M}_1$, $\mathcal{M}_2$, $\dots$, $\mathcal{M}_k$. 
$\mathcal{M}_i$ can be positive (Spherical), zero (Euclidean), negative (Hyperbolic) curvature space.
The product manifold is defined as the Cartesian product $\mathcal{M} = \mathcal{M}_1 \times \mathcal{M}_2 \times \dots \times \mathcal{M}_k$.  
We write a point $p \in \mathcal{M}$ through their coordinates $p=(p_1, \dots, p_k), p_i \in \mathcal{M}_i$. Similarly, a tangent vector $v \in T_p \mathcal{M}$ can be written as $(v_1, \dots , v_k) : v_i \in T_{p_i} \mathcal{M}_i$.
Gradient descent on manifolds requires the notion of taking steps.
This step can be performed in the tangent space and transferred to the manifold via the logarithmic map, and exponential map \cite{petersen2006riemannian}.
The product space is also equipped with a distance function. The squared distance between points $x, y \in  \mathcal{M}$ is defined as: $\quad d_{P}^{2}(x, y)=\sum_{i=1}^{k} d_{i}^{2}\left(x_{i}, y_{i}\right)$.

\section{Curvature estimation on graph data}
\label{theoretical-practical-analysis}

\paragraph{\textbf{Curvature estimation on simple graphs}}
There are three commonly used definitions for local graph curvature: Ollivier-Ricci \cite{ollivier2009ricci}, Forman-Ricci \cite{sreejith2016forman}, and sectional curvature \cite{gu2018learning}. 
In this paper, we use sectional curvature for estimating the geometric structures of graphs. 
Sectional curvature is determined by geometric triangle properties as follows.

\textbf{Theorem 1:} Recall from \cite{gu2018learning} that on a given constant curvature geometric space, if $abc$ is a geodesic triangle and $m$ is the midpoint of $bc$, then $d(a,m)^2+ \frac{d(b,c)^2}{4} - \frac{d(a,b)^2 + d(a,c)^2}{2}$ is equal to zero when the underlying space is Euclidean, is positive in spherical and negative in hyperbolic space, respectively.

\textit{Proof:}
\begin{figure}[!h]
    \centering
    \includegraphics[scale = 0.5]{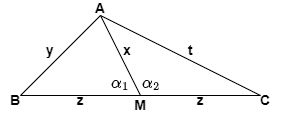}
    \caption{Geodesic triangle visualization.}
    \label{structure-information}
\end{figure}
 We provide proof of Theorem 1.
\begin{align}
A &= d(a,m)^2+ \frac{d(b,c)^2}{4} - \frac{d(a,b)^2 + d(a,c)^2}{2} \\
    &= x^2 + z^2 - \frac{y^2}{2} - \frac{t^2}{2} \\
&= \frac{1}{2} (2x^2 + 2z^2 - y^2 - t^2) \\
&= \frac{1}{2} [(x^2 +z^2 - y^2) + (x^2 + z^2 - t^2)] \\
&= \frac{1}{2} [2xz \operatorname{cos}{\alpha_1} + 2 xz \operatorname{cos}{\alpha_2}] \\
&= xz (\operatorname{cos}{\alpha_1} +  \operatorname{cos}{\alpha_2})
\end{align}

From Equation (6), we apply the cosine rule \footnote{https://en.wikipedia.org/wiki/Law\_of\_cosines}.
We have three cases:
\begin{itemize}
    \item $\operatorname{cos}\alpha_1 + \operatorname{cos} \alpha_2 = 0$: $\alpha_1$ and $\alpha_2$ are two supplementary angles, $\alpha_1 + \alpha_2 = 180^0$. Then the triangle is in Euclidean space.
     \item Similarly, it will be negative in hyperbolic and positive in the spherical curvature space.
\end{itemize} 

\paragraph{\textbf{Curvature estimation on graph data}}

Given theorem (1), let $v$ be a node in $G$; $b, c$ neighbors of $v$ and $a$ any other node. 
Then, the sectional curvature of a node $v$ and its neighbors $b,c$ is defined following: $\frac{1}{|V|-3} \sum_{a \in G \backslash\{v, b, c\}} \xi_G(v ; b, c ; a)$ where 
\begin{equation}
\resizebox{0.8\hsize}{!}{
    $\xi_G(v ; b, c ; a)=\frac{1}{2 d_G(a, v)}\left(d_G(a, v)^2+\frac{d_G(b, c)^2}{4}-\frac{d_G(a, b)^2+d_G(a, c)^2}{2}\right)$}
\end{equation}
and $2d_G(v;b,c)$ is included to yield the right scalings for trees and cycles.

Next, we estimate the curvature of some typical topology graph structures.
Star $\mathbf{S}_n$ is created from one central node and $n$ leaves. We consider $n \geq 3$, the local curvature at the center node $v$ with two neighbors $b, c$ is $-1$.
Tree $\mathbf{T}_b$ with branching factor $b$ is the finite depth tree with $b \geq 2$. The sectional curvature on the tree in the range $\xi(T) \in [-1, 0]$.
Cycles graph $\mathbf{C}_n$ with $n \geq 4$. If $n$ is even, then $\xi_{C_n}(v; b,c;a) = 0$ for all points except the one diametrically opposite to $v$ for which have $\xi_{C_n}(v; b,c;a) = 1$. 
If $n$ is odd, then for two points we have $\xi_{C_n}(v; b,c;a) = \frac{n}{2(n-1)}$.
As a result, $\xi(C_n) = \frac{1}{n-3}$ for even $n$ and $\xi(C_n) = \frac{n}{(n-1)(n-3)}$ for odd $n$.

\begin{figure*}[ht]
    \centering
    \includegraphics[scale = 0.45]{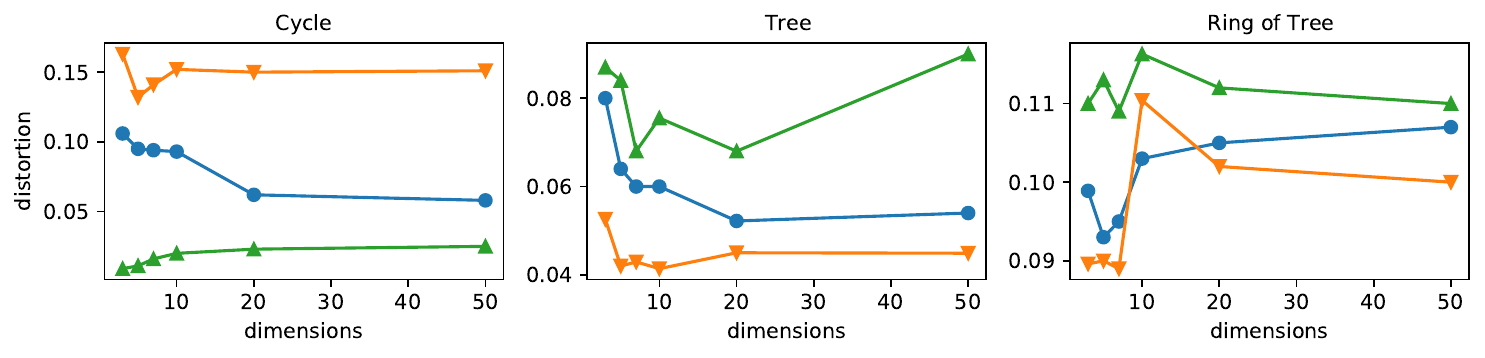}
    \caption{Comparison of using single spaces in Cycle and Tree data sets when gradually increasing the number of dimensions. Blue is Euclidean, Orange is Hyperbolic and Green is Spherical.}
    \label{distortion_cycle_tree_datasets}
\end{figure*}

\paragraph{\textbf{Distortion error on simple graphs}}
We have demonstrated the limitations of using a single curvature approach to embed graphs with varying topologies. 
To investigate the impact of curvature spaces on the quality of embedding spaces, we conducted experiments on three synthetic datasets with specific structures, including trees, circles, and rings of trees (Table \ref{tab:synthetic_statistic}). 
Figure \ref{distortion_cycle_tree_datasets} shows the distortion error results for Cycle and Tree graphs. 
Our findings suggest that different graph structures require corresponding curvature spaces for optimal embedding quality. 
For instance, spherical space (positive curvature) provides the least distortion error for cycle-like datasets (from $\mathbf{S}_3$ to $\mathbf{S}_{50}$), while hyperbolic spaces (negative curvature) give a minimal error for tree-like datasets (from $\mathbf{H}_3$ to $\mathbf{H}_{50}$).
All three models show some advancements compared to others in certain cases. 
However, the overall distortions achieved are significantly higher than when using hyperbolic space with tree-like or spherical space with circle-like data. 
For example, the distortion error on the Cycle tree is 0.09 compared to 0.02 on $\textbf{H}_{10}$ with Cycle data and 0.042 on $\textbf{S}_5$ with simple Tree data. 
Therefore, using a product of individual spaces can improve the accuracy of embedding data with a mixture of structures.
\begin{table}
\centering
\caption{ Statistics of all four synthetic datasets.}
\begin{tabular}{c c c}
\small
Dataset   & \#Nodes ($|V|$) & \#Edges ($|E|$)  \\ 
\hline
  Tree          &      40            &        39            \\
  Cycle          &       40           &         40           \\  
Circle of tree       &       40           &          40         \\
\hline
\end{tabular}
\label{tab:synthetic_statistic}
\end{table}

\section{Additional experimental results}

\subsection{Graph reconstruction task}
\label{apendix_graph_reconstruction}

\paragraph{\textbf{Datasets}}
The synthetic datasets we use are small graphs with 40 nodes that are designed to have specific geometric structures, including a circle, a tree, and a ring of trees. 
To assess the effectiveness of our approach on larger and more complex graphs, we also use three benchmark datasets: CsPhD \cite{de2011exploratory}, Power \cite{watts1998collective}, and Cities \cite{burkardt2011cities}. 
The Cities dataset consists of 1025 nodes and 1043 edges, while the Power dataset contains 4941 nodes and 6594 edges. Additionally, the CsPhD dataset has 312 nodes and 48516 edges. 

\paragraph{\textbf{Baselines}}
We compare the distortion error of node embeddings on both synthetic and benchmark datasets between our proposed model and the product spaces (PM) \cite{gu2018learning} method.

\paragraph{\textbf{Metrics}}
We use two standard metrics to measure the quality of embeddings: average distortion $D_{\text{avg}}$ and mean average precision $mAP$.
$D_{\text{avg}}$ is a global metric that considers all the exact distance values. 
Let $G = (V, E)$ be a graph and node $a \in V$ have a neighborhood $\mathcal{N}_a = {b_1, \cdots, b_{\text{deg}(a)} }$, where $\text{deg}(a)$ is the degree of $a$. 
In the embedding $f$, define $R_{a, b_i}$ to be the smallest ball around $f(a)$ that contains $b_i$, which means $R_{a, b_i}$ is the smallest set of nearest points required to retrieve the $i^\text{th}$ neighbor of $a$ in $f$. 
Thus, $mAP = \frac{1}{|V|} \sum\limits_{a \in V} \frac{1}{\text{deg}(a)} \sum\limits_{i = 1}^{\left|\mathcal{N}{a}\right|} \frac{\left|\mathcal{N}{a} \cap R_{a, b} \right|}{\left|R_{a, b_{i}}\right|}$. $mAP$ is a ranking-based measure for local neighborhoods, and it does not track exact distances like $D_{\text{avg}}$.

\begin{table}[ht]
\centering
\caption{Average distortion $D_\text{avg}(\downarrow)$ and mAP $(\uparrow)$ on CsPhDs and Power datasets on dimensions 10 and 50 respectively. The best $D_\text{avg}$ results are in boldface, while the second best results are underlined.}
\vspace{1mm}
\resizebox{0.8\columnwidth}{!}{%
\begin{tabular}{c c c c c c c}
\hline

\multicolumn{1}{c}{Method} &\multicolumn{1}{c}{Models} &  \multicolumn{2}{c}{Cs PhDs} &  \multicolumn{2}{c}{Power} \\ 
 && \multicolumn{1}{c}{$D_\text{avg}$} &\multicolumn{1}{c}{mAP} & \multicolumn{1}{c}{$D_\text{avg}$} &\multicolumn{1}{c}{mAP}\\

\hline
\small{PM} &$(\mathbb{H}^{2})^5$           & $\underline{0.0357}$ & 0.9694   & 0.0396 & 0.8739\\ 
&$(\mathbb{H}^{5})^2$          & 0.0382 & 0.9628  & 0.0365 & 0.8605 \\ 
&$(\mathbb{S}^{5})^2$           & 0.0579 & 0.7940 & 0.0471 & 0.8059 \\
&$(\mathbb{S}^{2})^5$           & 0.0562 & 0.8314 & 0.0483 & 0.8818\\
&$\mathbb{H}^{5} \times \mathbb{S}^{5}$          & 0.0529 & 0.9041  & $\underline{0.0323}$ & 0.8850 \\ 

\small{\ourmodel}                  &          \\
&$\mathbf{s}(\mathbb{H}^{2})^5$          & $\mathbf{0.0301}$ & 0.9699  & 0.0423 & 0.8540 \\ 
&         & \tiny{($+15.6\%$)}   \\ 

&$\mathbf{s}(\mathbb{H}^{5})^2$           & 0.0489 & 0.8465 & 0.034 & 0.88 \\ 
&$\mathbf{s}(\mathbb{S}^{5})^2$           & 0.056 & 0.813 & 0.0431 & 0.812 \\ 
&$\mathbf{s}(\mathbb{S}^{2})^5$          & 0.055 & 0.8361 & 0.0402 & 0.894 \\ 
&$\mathbf{s}_1\mathbb{H}^{5} \times \mathbf{s}_2\mathbb{S}^{5}$           & 0.0494 & 0.9231 & $\mathbf{0.0231}$ & 0.8842 \\ 
&& &           & \tiny{($+28.4\%$)}    \\
\hline
\small{PM}                  &          \\
&$(\mathbb{H}^{10})^5$          & 0.0657 & 0.913   &  0.0501 & 0.869  \\ 
&$(\mathbb{H}^{5})^{10}$          & 0.0786 & 0.8600   & 0.0723 & 0.842 \\ 
&$(\mathbb{S}^{10})^5$           & 0.0726 & 0.843  & 0.0682 & 0.84 \\ 
&$(\mathbb{S}^{5})^{10}$          & 0.07 & 0.89  & 0.0701 & 0.831\\ 
&$(\mathbb{H}^{10})^3 \times (\mathbb{S}^{10})^2 $          & 0.0620 & 0.9120  & 0.0510 & 0.8500 \\ 
&$(\mathbb{H}^{10})^2 \times (\mathbb{S}^{10})^2 \times \mathbb{E}^{10}$          & $\underline{0.0571}$ & 0.9071  & $\underline{0.0495}$ & 0.8561  \\ 

\small{\ourmodel}                  &          \\
&$\mathbf{s}(\mathbb{H}^{10})^5$          & 0.062 & 0.932    & 0.0479 & 0.871\\ 
&$\mathbf{s}(\mathbb{H}^{5})^{10}$          & 0.0591 & 0.8900  & 0.0503 & 0.8570\\ 
&$\mathbf{s}(\mathbb{S}^{10})^5$           & 0.074 & 0.829  & 0.0613 & 0.86\\ 
&$\mathbf{s}(\mathbb{S}^{5})^{10}$           & 0.062 & 0.87  & 0.0648 & 0.871\\ 

&$\mathbf{s_1}(\mathbb{H}^{10})^3 \times \mathbf{s_2}(\mathbb{S}^{10})^2 $          & 0.0590 & 0.9312  & $\mathbf{0.0426}$ & 0.8620 \\ 
& & &  & \tiny{($+13.9 \%$)}  \\ 

&$\mathbf{s_1}(\mathbb{H}^{10})^2 \times \mathbf{s_2}(\mathbb{S}^{10})^2 \times \textbf{s}_3 \mathbb{E}^{10}$          & $\mathbf{0.0460}$ & 0.9240  & 0.0502 & 0.8600\\ 
&  & \tiny{($+19.3 \%$)}  \\

\hline
\end{tabular}}
\label{tab:benchmark_csPhD_datasets_full}
\end{table}

\subsection{Additional information for Recommendation task}
\label{apendix_rec_sys}
\paragraph{\textbf{Metrics}}
We use two measures \textit{Hit Ratio} (H) \cite{deshpande2004item} and \textit{Normalized Discounted Cumulative Gain (N)} \cite{he2015trirank} to examine the predictive ability of these models. 
The final $H@k$ and $N@k$ are averaged on all users' $H@k$ and $N@k$ scores. 
We choose $k = 10$ to evaluate the model. 

\paragraph{\textbf{Datasets}} We perform experiments on two popular datasets, MovieLens-1M and LastFM-20K. The LastFm dataset \citep{Bertin-Mahieux2011} is obtained from a music website\footnote{http://millionsongdataset.com/lastfm/}. It is preprocessed to have 1892 users and 17632 music records. The MovieLens-1M is created from 6040 users and 3706 movies.

\paragraph{\textbf{Baselines}}
We consider the three works below as the baselines for our model:  CML \cite{hsieh2017collaborative}, HyperML \cite{vinh2020hyperml}.
For specific, CML \cite{hsieh2017collaborative} investigates the relationship between metric learning and collaborative filtering. 
It proposes a method that learns a joint metric space capable of encoding not only users' preferences but also the similarity between users and items.
HyperML \cite{vinh2020hyperml} presents the connection between metric learning in hyperbolic space and collaborative filtering by exploring hyperbolic geometry.
HyperML-PM is our extension of HyperML in the product of model space.
HyperML-WPM (Our) is our extension of HyperML in the product of model spaces with the gating mechanism.

\subsection{Additional information for Knowledge graph embedding}
\label{apendix_kg_embedding}
\paragraph{\textbf{Metrics}}
The performance of various models is evaluated using two standard metrics: mean reciprocal rank (MRR) and hit rate (HR@3).

\paragraph{\textbf{Datasets}}
We used two standard datasets, WN18RR \cite{dettmers2018convolutional} and FB15K-237 \cite{bordes2013translating}, for our analysis. WN18RR is derived from WordNet, a lexical database of semantic relations between words. FB15K-237 is a subset of the Freebase knowledge graph, which is a comprehensive resource containing general information. 
Table \ref{tab:knowledge_graph_statistic} shows the statistics of the two datasets.

\begin{table}
\centering
\caption{ Statistics the number of entities, relations, and training, validations, and test triplets for two datasets.}
\vspace{1mm}
\resizebox{0.7\columnwidth}{!}{%
\begin{tabular}{c c c c c c}
\small
Dataset   & \#entity & \#relation & \#training & \#validation & \#test  \\ 
\hline
  FB15k-237          &   14,541      &     1,345     & 272,115    & 17,535   & 20,466          \\
  WN18RR          &       40,943           &     11 & 86,835 & 3,034 & 3,134           \\  
\hline
\end{tabular}}
\label{tab:knowledge_graph_statistic}
\end{table}

\paragraph{\textbf{Objective function}}
Given a knowledge graph $\mathcal{G}$ with a set of entities $\mathcal{E}$ and a set of relation $\mathcal{R}$. Each triplet $(h,r,t) \in \mathcal{G}$ is included by head entity $h$, tail entity $t$, and the relation $r \in \mathcal{R}$ between them. 

There are a lot of works that propose RotE \cite{sun2019rotate} in Euclidean space, and RotH \cite{chami2019low} in Hyperbolic space. In this work, we extend to the product of different curvature spaces. Formally, entities $h, t$ are represented by vector $\mathbf{e}_h, \mathbf{e}_t \in \mathbb{R}^b$ and the relation $r$ is represented by two translation vectors $\alpha_r, \beta_r \in \mathbb{R}^b$ and a rotation vector $\gamma_r \in \mathbb{R}^b$. The head entity will translate twice via Mobius addition operation and rotate one time.
\begin{align}
    Q(h,r)= \operatorname{Rot}(\operatorname{exp}_0^c(\mathbf{e}_h) \oplus_c \operatorname{exp}_0^c(\alpha_r), \gamma_r) \oplus_c \operatorname{exp}_0^c (\beta_r)
\end{align}
with $c > 0$ and $\operatorname{exp}_0^c$ is the exponential map over the origin. $\operatorname{Rot}$ is a rotation function with $\gamma_r$ is the rotation matrix.

According to the above definition, for each triple $(h,r,t)$, we define the distance function as:
\begin{align}
    d_r({h}, {t}) = \sqrt{d_{\mathcal{M}_c}^2 (Q({h},r), \operatorname{exp}_0^c(\textbf{e}_t))}
\end{align}
where $\mathcal{M}_c$ is the product of curvature manifold. In \cite{sun2019rotate}, the distance function of RotatE for the triple $(h,r,t)$ is defined as: $d_r({h}, {r}) = || {h} \odot {r} - {t}||$

The final negative sampling loss is defined by the cross-entropy loss:
\begin{align}
    \mathcal{L} =\sum_{(h,r,t) \in \Omega} \operatorname{log}(1+ \operatorname{exp}(-Y_{(h,r,t)} d_r(h,t)))
\end{align}

where $Y_{(h,r,t)} \in \{1, -1\}$ is a binary label indicating whether a triplet is real or not.

\paragraph{\textbf{Baselines}}
RotatE \cite{sun2019rotate} is a knowledge graph embedding that is used to learn the representations of entities and relations in knowledge graphs.
RotatH is the extension of RotatE \cite{sun2019rotate} in the hyperbolic space.
Product-RotatH is the extension of RotatE in the product of the hyperbolic spaces \cite{takeuchi2022neural}.
SwisE \cite{zhang2021switch} used the gating mechanism which is learned to choose the component space for knowledge graph embedding.
\ourmodel{}-Rotat is our extension by using the product of manifold in representing the relations among entities in the knowledge graph.

\subsection{Additional information for Node Classification and Link Prediction}
\label{apendix_node_graph_cls}
\paragraph{\textbf{Metrics}} We utilize ROC AUC as a metric to evaluate the performance of Link Prediction (LP), whereas we rely on the F1 score to assess the Node Classification (NC) performance. In both cases, a higher score indicates better performance.

\paragraph{\textbf{Datasets}} In this experiment, we evaluate model performance on the two different benchmark datasets.
DISEASE is the dataset of Infectious diseases from Oxford University \cite{anderson1992infectious}.
AIRPORT: is the dataset of airline routes from OpenFlight.org. Each node represents an airport, and the edge represents airline routes among these airports.
Detailed information regarding these datasets is provided in Table \ref{gcn-benmark-dataset-statistic}.
 
\begin{table}
\caption{Benchmark's statistic for Node Classification and Link Prediction tasks.}
\vspace{1mm}
\centering

\begin{tabular}{c c c c c c}
Dataset   & \#Nodes & \#Edges & \#Classes & $\delta$ \\ 
\hline
  DISEASE          &      1044           &        1043 & 2     & 0      \\
  AIRPORT          &       3188           &         18631     & 4 & 1  \\  
\hline
\end{tabular}
\label{gcn-benmark-dataset-statistic}
\end{table}

\paragraph{\textbf{Baselines}} We evaluate the contributions of our proposed model by measuring the F1 and AUC scores on two datasets, compared with five different baseline models:
MLP and Hyperbolic-MLP are two variants of multilayer perceptron (MLP) classifiers operating on the Euclidean ($\mathbf{E}$) and hyperbolic space ($\mathbf{H}$), respectively.
HGCN \cite{chami2019hyperbolic} is an extension of graph convolutional networks (GCNs) to hyperbolic geometry.
Product-HGCN \cite{takeuchi2022neural} extends GCNs in the product of hyperbolic geometries.
Mix-GCN \cite{gu2018learning} extends GCNs in the product of hyperbolic, spherical, and Euclidean spaces.
Our proposed model (\ourmodel-GCN) extends GCNs with a gating mechanism in the product of different curvature spaces (H, E, S).

\end{document}

%% file: math_commands.tex
\usepackage{amsmath,amsfonts,bm}

\def\eqref#1{equation~\ref{#1}}

\def\1{\bm{1}}

\DeclareMathAlphabet{\mathsfit}{\encodingdefault}{\sfdefault}{m}{sl}
\SetMathAlphabet{\mathsfit}{bold}{\encodingdefault}{\sfdefault}{bx}{n}













%% file: main.bbl
\begin{thebibliography}{43}
\providecommand{\natexlab}[1]{#1}
\providecommand{\url}[1]{\texttt{#1}}
\expandafter\ifx\csname urlstyle\endcsname\relax
  \providecommand{\doi}[1]{doi: #1}\else
  \providecommand{\doi}{doi: \begingroup \urlstyle{rm}\Url}\fi

\bibitem[Anderson \& May(1992)Anderson and May]{anderson1992infectious}
Roy~M Anderson and Robert~M May.
\newblock \emph{Infectious diseases of humans: dynamics and control}.
\newblock Oxford university press, 1992.

\bibitem[Bengio et~al.(2013)Bengio, Courville, and
  Vincent]{bengio2013representation}
Yoshua Bengio, Aaron Courville, and Pascal Vincent.
\newblock Representation learning: A review and new perspectives.
\newblock \emph{IEEE transactions on pattern analysis and machine
  intelligence}, 2013.

\bibitem[Bertin-Mahieux et~al.(2011)Bertin-Mahieux, Ellis, Whitman, and
  Lamere]{Bertin-Mahieux2011}
Thierry Bertin-Mahieux, Daniel~P.W. Ellis, Brian Whitman, and Paul Lamere.
\newblock The million song dataset.
\newblock In \emph{{Proceedings of the 12th International Conference on Music
  Information Retrieval}}, 2011.

\bibitem[Bordes et~al.(2013)Bordes, Usunier, Garcia-Duran, Weston, and
  Yakhnenko]{bordes2013translating}
Antoine Bordes, Nicolas Usunier, Alberto Garcia-Duran, Jason Weston, and Oksana
  Yakhnenko.
\newblock Translating embeddings for modeling multi-relational data.
\newblock \emph{Advances in neural information processing systems}, 2013.

\bibitem[Bridson \& Haefliger(2013)Bridson and Haefliger]{bridson2013metric}
Martin~R Bridson and Andr{\'e} Haefliger.
\newblock \emph{Metric spaces of non-positive curvature}, volume 319.
\newblock Springer Science \& Business Media, 2013.

\bibitem[Burkardt(2011)]{burkardt2011cities}
John Burkardt.
\newblock Cities--city distance datasets.
\newblock \emph{URL https://rdrr. io/cran/TSP/man/USCA312. html}, 2011.

\bibitem[Cai et~al.(2021)Cai, Wang, and Wang]{cai2021graph}
Chen Cai, Dingkang Wang, and Yusu Wang.
\newblock Graph coarsening with neural networks.
\newblock In \emph{International conference on machine learning}, 2021.

\bibitem[Cao et~al.(2022)Cao, Xu, Yang, Cao, and Huang]{cao2022geometry}
Zongsheng Cao, Qianqian Xu, Zhiyong Yang, Xiaochun Cao, and Qingming Huang.
\newblock Geometry interaction knowledge graph embeddings.
\newblock In \emph{Proceedings of the AAAI Conference on Artificial
  Intelligence}, volume~36, 2022.

\bibitem[Chami et~al.(2019)Chami, Ying, R{\'e}, and
  Leskovec]{chami2019hyperbolic}
Ines Chami, Zhitao Ying, Christopher R{\'e}, and Jure Leskovec.
\newblock Hyperbolic graph convolutional neural networks.
\newblock \emph{Advances in neural information processing systems}, 2019.

\bibitem[Chami et~al.(2020{\natexlab{a}})Chami, Gu, Chatziafratis, and
  R{\'e}]{chami2020trees}
Ines Chami, Albert Gu, Vaggos Chatziafratis, and Christopher R{\'e}.
\newblock From trees to continuous embeddings and back: Hyperbolic hierarchical
  clustering.
\newblock \emph{Advances in Neural Information Processing Systems},
  2020{\natexlab{a}}.

\bibitem[Chami et~al.(2020{\natexlab{b}})Chami, Wolf, Juan, Sala, Ravi, and
  R{\'e}]{chami2019low}
Ines Chami, Adva Wolf, Da-Cheng Juan, Frederic Sala, Sujith Ravi, and
  Christopher R{\'e}.
\newblock Low-dimensional hyperbolic knowledge graph embeddings.
\newblock In \emph{Proceedings of the 58th Annual Meeting of the Association
  for Computational Linguistics}, pp.\  6901--6914. Association for
  Computational Linguistics, 2020{\natexlab{b}}.

\bibitem[Chami et~al.(2020{\natexlab{c}})Chami, Wolf, Juan, Sala, Ravi, and
  R{\'e}]{chami2020low}
Ines Chami, Adva Wolf, Da-Cheng Juan, Frederic Sala, Sujith Ravi, and
  Christopher R{\'e}.
\newblock Low-dimensional hyperbolic knowledge graph embeddings.
\newblock \emph{Association for Computational Linguistics}, 2020{\natexlab{c}}.

\bibitem[de~Nooy et~al.(2011)de~Nooy, Mrvar, and Batagelj]{de2011exploratory}
Wouter de~Nooy, Andrej Mrvar, and Vladimir Batagelj.
\newblock Exploratory social network analysis with pajek: Revised and expanded
  second edition.
\newblock \emph{Cambridge: Cambridge University Press}, 2011.

\bibitem[Deshpande \& Karypis(2004)Deshpande and Karypis]{deshpande2004item}
Mukund Deshpande and George Karypis.
\newblock Item-based top-n recommendation algorithms.
\newblock \emph{ACM Transactions on Information Systems}, 2004.

\bibitem[Dettmers et~al.(2018)Dettmers, Minervini, Stenetorp, and
  Riedel]{dettmers2018convolutional}
Tim Dettmers, Pasquale Minervini, Pontus Stenetorp, and Sebastian Riedel.
\newblock Convolutional 2d knowledge graph embeddings.
\newblock In \emph{Proceedings of the AAAI conference on artificial
  intelligence}, 2018.

\bibitem[Di~Giovanni et~al.(2022)Di~Giovanni, Luise, and
  Bronstein]{di2022heterogeneous}
Francesco Di~Giovanni, Giulia Luise, and Michael Bronstein.
\newblock Heterogeneous manifolds for curvature-aware graph embedding.
\newblock \emph{ICLR Workshop on Geometrical and Topological Representation
  Learning}, 2022.

\bibitem[Finkelstein et~al.(2001)Finkelstein, Gabrilovich, Matias, Rivlin,
  Solan, Wolfman, and Ruppin]{finkelstein2001placing}
Lev Finkelstein, Evgeniy Gabrilovich, Yossi Matias, Ehud Rivlin, Zach Solan,
  Gadi Wolfman, and Eytan Ruppin.
\newblock Placing search in context: The concept revisited.
\newblock In \emph{Proceedings of the 10th international conference on World
  Wide Web}, 2001.

\bibitem[Gu et~al.(2018)Gu, Sala, Gunel, and R{\'e}]{gu2018learning}
Albert Gu, Frederic Sala, Beliz Gunel, and Christopher R{\'e}.
\newblock Learning mixed-curvature representations in product spaces.
\newblock In \emph{International Conference on Learning Representations}, 2018.

\bibitem[He et~al.(2015)He, Chen, Kan, and Chen]{he2015trirank}
Xiangnan He, Tao Chen, Min-Yen Kan, and Xiao Chen.
\newblock Trirank: Review-aware explainable recommendation by modeling aspects.
\newblock In \emph{Proceedings of the 24th ACM International on Conference on
  Information and Knowledge Management}, 2015.

\bibitem[Hsieh et~al.(2017)Hsieh, Yang, Cui, Lin, Belongie, and
  Estrin]{hsieh2017collaborative}
Cheng-Kang Hsieh, Longqi Yang, Yin Cui, Tsung-Yi Lin, Serge Belongie, and
  Deborah Estrin.
\newblock Collaborative metric learning.
\newblock In \emph{Proceedings of the 26th international conference on world
  wide web}, 2017.

\bibitem[Ji et~al.(2021)Ji, Pan, Cambria, Marttinen, and Philip]{ji2021survey}
Shaoxiong Ji, Shirui Pan, Erik Cambria, Pekka Marttinen, and S~Yu Philip.
\newblock A survey on knowledge graphs: Representation, acquisition, and
  applications.
\newblock \emph{IEEE transactions on neural networks and learning systems},
  2021.

\bibitem[Leimeister \& Wilson(2018)Leimeister and Wilson]{leimeister2018skip}
Matthias Leimeister and Benjamin~J Wilson.
\newblock Skip-gram word embeddings in hyperbolic space.
\newblock \emph{arXiv preprint arXiv:1809.01498}, 2018.

\bibitem[Li et~al.(2022)Li, Fu, Sun, Ji, Tan, Wu, and Peng]{curvaGAN}
Jianxin Li, Xingcheng Fu, Qingyun Sun, Cheng Ji, Jiajun Tan, Jia Wu, and Hao
  Peng.
\newblock Curvature graph generative adversarial networks.
\newblock 2022.

\bibitem[Liu et~al.(2019)Liu, Nickel, and Kiela]{liu2019hyperbolic}
Qi~Liu, Maximilian Nickel, and Douwe Kiela.
\newblock Hyperbolic graph neural networks.
\newblock \emph{Advances in Neural Information Processing Systems}, 2019.

\bibitem[Meng et~al.(2019)Meng, Huang, Wang, Zhang, Zhuang, Kaplan, and
  Han]{meng2019spherical}
Yu~Meng, Jiaxin Huang, Guangyuan Wang, Chao Zhang, Honglei Zhuang, Lance
  Kaplan, and Jiawei Han.
\newblock Spherical text embedding.
\newblock \emph{Advances in Neural Information Processing Systems}, 2019.

\bibitem[Mikolov et~al.(2013)Mikolov, Chen, Corrado, and
  Dean]{mikolov2013efficient}
Tomas Mikolov, Kai Chen, Greg Corrado, and Jeffrey Dean.
\newblock Efficient estimation of word representations in vector space.
\newblock \emph{International Conference on Learning Representations}, 2013.

\bibitem[Nickel \& Kiela(2017)Nickel and Kiela]{nickel2017poincare}
Maximillian Nickel and Douwe Kiela.
\newblock Poincar{\'e} embeddings for learning hierarchical representations.
\newblock \emph{Advances in neural information processing systems}, 2017.

\bibitem[Ollivier(2009)]{ollivier2009ricci}
Yann Ollivier.
\newblock Ricci curvature of markov chains on metric spaces.
\newblock \emph{Journal of Functional Analysis}, 2009.

\bibitem[Petersen(2006)]{petersen2006riemannian}
Peter Petersen.
\newblock \emph{Riemannian geometry}.
\newblock Springer, 2006.

\bibitem[Skopek et~al.(2020)Skopek, Ganea, and B{\'e}cigneul]{skopek2019mixed}
Ondrej Skopek, Octavian-Eugen Ganea, and Gary B{\'e}cigneul.
\newblock Mixed-curvature variational autoencoders.
\newblock In \emph{International Conference on Learning Representations}, 2020.

\bibitem[Sreejith et~al.(2016)Sreejith, Mohanraj, Jost, Saucan, and
  Samal]{sreejith2016forman}
RP~Sreejith, Karthikeyan Mohanraj, J{\"u}rgen Jost, Emil Saucan, and Areejit
  Samal.
\newblock Forman curvature for complex networks.
\newblock \emph{Journal of Statistical Mechanics: Theory and Experiment}, 2016.

\bibitem[Sun et~al.(2019)Sun, Deng, Nie, and Tang]{sun2019rotate}
Zhiqing Sun, Zhi-Hong Deng, Jian-Yun Nie, and Jian Tang.
\newblock Rotate: Knowledge graph embedding by relational rotation in complex
  space.
\newblock \emph{International Conference on Learning Representations}, 2019.

\bibitem[Takeuchi et~al.(2022)Takeuchi, Nishida, and
  Nakayama]{takeuchi2022neural}
Jun Takeuchi, Noriki Nishida, and Hideki Nakayama.
\newblock Neural networks in a product of hyperbolic spaces.
\newblock In \emph{Proceedings of the 2022 Conference of the North American
  Chapter of the Association for Computational Linguistics: Human Language
  Technologies}, 2022.

\bibitem[Topping et~al.(2022)Topping, Di~Giovanni, Chamberlain, Dong, and
  Bronstein]{topping2021understanding}
Jake Topping, Francesco Di~Giovanni, Benjamin~Paul Chamberlain, Xiaowen Dong,
  and Michael~M Bronstein.
\newblock Understanding over-squashing and bottlenecks on graphs via curvature.
\newblock In \emph{International conference on machine learning}, 2022.

\bibitem[Vinh~Tran et~al.(2020)Vinh~Tran, Tay, Zhang, Cong, and
  Li]{vinh2020hyperml}
Lucas Vinh~Tran, Yi~Tay, Shuai Zhang, Gao Cong, and Xiaoli Li.
\newblock Hyperml: A boosting metric learning approach in hyperbolic space for
  recommender systems.
\newblock In \emph{Proceedings of the 13th International Conference on Web
  Search and Data Mining}, 2020.

\bibitem[Watts \& Strogatz(1998)Watts and Strogatz]{watts1998collective}
Duncan~J Watts and Steven~H Strogatz.
\newblock Collective dynamics of ‘small-world’networks.
\newblock \emph{nature}, 1998.

\bibitem[Wilson et~al.(2014)Wilson, Hancock, Pekalska, and
  Duin]{wilson2014spherical}
Richard~C Wilson, Edwin~R Hancock, El{\.z}bieta Pekalska, and Robert~PW Duin.
\newblock Spherical and hyperbolic embeddings of data.
\newblock \emph{IEEE transactions on pattern analysis and machine
  intelligence}, 2014.

\bibitem[Xu et~al.(2022{\natexlab{a}})Xu, Wang, Chen, Lu, Duan, Zhou,
  et~al.]{xu2022hyperminer}
Yi~Xu, Dongsheng Wang, Bo~Chen, Ruiying Lu, Zhibin Duan, Mingyuan Zhou, et~al.
\newblock Hyperminer: Topic taxonomy mining with hyperbolic embedding.
\newblock \emph{Advances in Neural Information Processing Systems}, 35,
  2022{\natexlab{a}}.

\bibitem[Xu et~al.(2022{\natexlab{b}})Xu, Wen, Wang, Liu, Wang, Yang, Ding,
  Zhang, Zhang, Xu, et~al.]{xu2022amcad}
Zhirong Xu, Shiyang Wen, Junshan Wang, Guojun Liu, Liang Wang, Zhi Yang, Lei
  Ding, Yan Zhang, Di~Zhang, Jian Xu, et~al.
\newblock Amcad: Adaptive mixed-curvature representation based advertisement
  retrieval system.
\newblock In \emph{2022 IEEE 38th International Conference on Data Engineering
  (ICDE)}, pp.\  3439--3452. IEEE, 2022{\natexlab{b}}.

\bibitem[Ying et~al.(2018)Ying, You, Morris, Ren, Hamilton, and
  Leskovec]{ying2018hierarchical}
Zhitao Ying, Jiaxuan You, Christopher Morris, Xiang Ren, Will Hamilton, and
  Jure Leskovec.
\newblock Hierarchical graph representation learning with differentiable
  pooling.
\newblock \emph{Advances in neural information processing systems}, 2018.

\bibitem[You et~al.(2022)You, Tran, and Lee]{you2022multi}
Di~You, Thanh Tran, and Kyumin Lee.
\newblock Multi-behavior recommendation with hyperbolic geometry.
\newblock In \emph{2022 IEEE International Conference on Big Data (Big Data)},
  pp.\  1750--1759. IEEE, 2022.

\bibitem[Yu \& De~Sa(2023)Yu and De~Sa]{yu2023random}
Tao Yu and Christopher De~Sa.
\newblock Random laplacian features for learning with hyperbolic space.
\newblock In \emph{International Conference on Learning Representations}, 2023.

\bibitem[Zhang et~al.(2021)Zhang, Tay, Jiang, Juan, and Zhang]{zhang2021switch}
Shuai Zhang, Yi~Tay, Wenqi Jiang, Da-cheng Juan, and Ce~Zhang.
\newblock Switch spaces: Learning product spaces with sparse gating.
\newblock \emph{arXiv preprint arXiv:2102.08688}, 2021.

\end{thebibliography}
